\documentclass[letterpaper, 10 pt, conference]{ieeeconf}  % Comment this line out if you need a4paper

\IEEEoverridecommandlockouts                              % This command is only needed if 
\usepackage{algorithm}% http://ctan.org/pkg/algorithms
\usepackage{algpseudocode}% http://ctan.org/pkg/algorithmicx
\usepackage{graphics}
\usepackage{cite}
\usepackage{amsmath,amssymb,amsfonts}
\usepackage{subfig}
\usepackage{textcomp}
\usepackage{xcolor}
\usepackage{nicefrac}       % compact symbols for 1/2, etc.
\usepackage[pdftex]{graphicx}
\usepackage{multirow}
\usepackage{xspace}
\usepackage{booktabs}
\usepackage{tabu}
\usepackage{siunitx}
\usepackage{authblk}

\title{\LARGE \bf
Feedback Control for Online Training of Neural Networks
}

\author{Zilong Zhao$^{1}$, Sophie Cerf$^{1}$ , Bogdan Robu$^{1}$ and Nicolas Marchand$^{1}$% <-this % stops a space
	\thanks{$^{1}$ GIPSA-lab, Univ. Grenoble Alpes, CNRS, Grenoble INP, 38000 Grenoble, France
		{\tt\small firstname.lastname@gipsa-lab.fr}}%
}

\makeatletter
\def\endthebibliography{%
  \def\@noitemerr{\@latex@warning{Empty `thebibliography' environment}}%
  \endlist
}
\makeatother
\newcolumntype{L}[1]{>{\raggedright\let\newline\\\arraybackslash\hspace{0pt}}m{#1}}
\newcolumntype{C}[1]{>{\centering\let\newline\\\arraybackslash\hspace{0pt}}m{#1}}
\newcolumntype{R}[1]{>{\raggedleft\let\newline\\\arraybackslash\hspace{0pt}}m{#1}}
\begin{document}
\maketitle
%\thispagestyle{empty}
%\pagestyle{empty}

%%%%%%%%%%%%%%%%%%%%%%%%%%%%%%%%%%%%%%%%%%%%%%%%%%%%%%%%%%%%%%%%%%%%%%%%%%%%%%%%
\begin{abstract}
Convolutional neural networks (CNNs) are commonly used for image classification tasks, raising the challenge of their application on data flows. During their training, adaptation is often performed by tuning the learning rate. Usual learning rate strategies are time-based i.e. monotonously decreasing. In this paper, we advocate switching to a performance-based adaptation, in order to improve the learning efficiency. We present E (Exponential)/PD (Proportional Derivative)-Control, a conditional learning rate strategy that combines a feedback PD controller based on the CNN loss function, with an exponential control signal to smartly boost the learning and adapt the PD parameters. Stability proof is provided as well as an experimental evaluation using two state of the art image datasets (CIFAR-10 and Fashion-MNIST). Results show better performances than the related works (faster network accuracy growth reaching higher levels) and robustness of the E/PD-Control regarding its parametrization.
\end{abstract}

%%%%%%%%%%%%%%%%%%%%%%%%%%%%%%%%%%%%%%%%%%%%%%%%%%%%%%%%%%%%%%%%%%%%%%%%%%%%%%%%
\section{Introduction}
\label{sec:Introduction}

Convolutional neural networks (CNNs) are popular machine learning algorithms for image classification , as they are well suited for visual pattern recognition and require low preprocessing \cite{Lecun2015}. Like all neural networks, CNNs are parametrized with so-called weights that enable to tune the network prediction model to fit the task. Those weights are learned iteratively based on training data using methods such as gradient descent. In this paper, we consider a scenario where data comes dynamically in batches (not all data is initially available), previous data batches being discarded at the arrival of a new one. This type of scenarios is very common in our everyday life if we think about sequential collection of a video flow or daily crowdsourcing \cite{4534830}\cite{Lease2011OnQC}. %\sophie{put 1 or 2 refs of sequential collection of video flows or crowd sourcing}. 

The learning rate parameter is used to weight the impact of a new epoch on the previously learned model. Thus, in the gradient-based algorithms, there are two factors that influence the reach of the gradient global minimum: the network's weights initialization and the learning rate policy. The weights initialization is often dealt with by setting them all null or generated from a uniform distribution \cite{DBLP:journals/jmlr/GlorotB10}.
The learning rate controls the speed to approach the minimum. A large learning rate will accelerate the converging speed but at the risk of diverging \cite{Bengio2012}. A small learning rate will slowly approach the minimum with less tendency to skip over it, but may fall into a local minimum. 

The objective is thus to set the learning rate strategy in order to learn from the data as fast as possible to reach the maximum CNN's predictions accuracy. The dynamic data collection scenario raises the challenge of a learning rate scheduling able to deal with the combination of epoch learning (take the most out of the currently available data) with batch learning (being able to include new data without forgetting the previous ones).

A learning rate strategy is defined by its initial value and its evolution law. The tuning of both is a significant challenge for the deep learning community \cite{Ioffe2015}. According to Bengio \cite{Bengio2012}, a learning rate of $0.01$ typically works as a default value for standard multi-layer neural networks. He also recommends a classic strategy to find a more suitable value for a given architecture and dataset. Its principle is to try several values on a subset of the dataset and compare the best validation accuracy for a fixed training time; and the lowest training time to reach a given validation accuracy \cite{Smith2017}. %This technique, similar to a static characterization, has the drawback of requiring an initial off-line phase and it supposes that the dynamic of data coming in is constant.
Learning rate evolution laws are usually of two kinds: time-based or adaptive.
%Usual learning rate evaluation laws are time based \cite{Bengio2012}.
Time-based learning strategies \cite{Bengio2012} are the most famous ones: the learning rate follows a predefined function (polynomial, exponential, etc.) that should decrease through time to ensure stability. 
The adaptive techniques are based on the gradient: they are reactive techniques that set the learning rate according to the past values of the gradient \cite{Ruder2016,Wu2018}. Given the definition of the gradient descent training law, a learning rate indexed on the gradient value is always decreasing with time.
The monotonous decrease of the learning rate is a well-established use that has rarely been questioned. However, Smith \cite{Smith2017} presents promising results with a cyclical learning rate law of triangular shape, and An et al. \cite{W8305126} introduced a time decreasing law with small sine oscillations. Indeed, we advocate that a brief increase in the learning rate could enable both to reach faster the global minimum and avoid being blocked in a local one.

% TO PUT IN EVAL: Indeed, having a small learning rate in the initial moments enable to find the right direction to minimize the loss. Afterwards, going fast in this direction (larger learning rate) avoids local minima that could be in the way. Eventually, a small learing rate enables to refine the minima when we found we reached one. %This strategy should be repeated at each new batch but with a scaling process to

Moreover, the issue with the state of the art learning rate laws regarding our continual learning scenario is that they do not take into account the dynamic of data coming in. They are predefined functions that do not adapt to the performances of the CNN training. 
Even by re-initializing the learning rate rule at each batch, it will not take into account the precision improvement through time that results from the memory of the previous batches.

In this work, we advocate using a control-based approach to adapt the learning rate in order to reach a high network accuracy in a short amount of time. % based on the network prediction performances. 
The principle is to switch from time decreasing rules to a performance-based rule to be able to increase the learning rate when necessary. 
%Loss functions, such as cross-entropy, are commonly used to evaluate how well an algorithm models a given dataset: it is our output signal.  
P and PD control strategies are initially developed. Then we present E/PD-Control, a hybrid strategy for setting the learning rate that combines both a time-based rule, with a first initial phase of an exponential growth of the learning rate, with a PD controller triggered by the network loss function. The initial E phase additionally allows the PD to be tuned on-line, thus getting rid of the need of an off-line profiling phase to adapt to a new dataset or network architecture. %The PI parameters are adapted thanks to the initial phase, and thus require no tuning. 
The E/PD-Control is evaluated on two classical state of the art datasets (CIFAR-10\cite{Krizhevsky09} and Fashion-MNIST\cite{xiao2017/online}), which are labelled image datasets commonly used to train computer vision algorithms\cite{He2016DeepRL}. Our control shows higher accuracy, faster rising time, lower final loss and more stable results than the state of the art techniques. Robustness regarding the initial value of the learning rate is also illustrated. 

In the remaining of the paper, we first present the problem statement in a control theory formulation (Section \ref{sec:Background}) and illustrate two state of the art learning rate strategies (Section \ref{sec:Motivation}). The control law is presented in Section \ref{sec:Control} and its stability analysis and performance evaluation are given in Section \ref{sec:Evaluation}.

\section{Background}
\label{sec:Background}

This section presents, in control terms, the system we aim at monitoring, its disturbances, the signals that evaluate its performances and the available control knob.  

	\subsection{The plant: a convolutional neural network}
Convolutional Neural Networks (CNN) are state of the art learning mechanisms that give the best results comparing to other learning algorithms when modeling image datasets \cite{Yamashita2018}. CNNs are inspired by the organization of animal visual cortex~\cite{hubel:monkey,fukushima:neocognitronbc}. They are a type of deep learning model made to process data that have grid patterns (neighboring features form a local structure such as in images). They are designed to automatically and adaptively learn spatial hierarchies of features, from low to high-level patterns. 
CNN training is done using algorithms  such as the Stochastic Gradient Descent (SGD) optimizer. We use SGD as it enables to set the learning rate at the beginning of each training epoch.
%According to their complexity, different tasks require different CNN configuration. 
%Networks architecture optimization is a key challenge in deep learning, which however is outside of the scope of this work. 

\begin{figure}[!hb]
	\begin{center}
		\includegraphics[width=0.5\columnwidth]{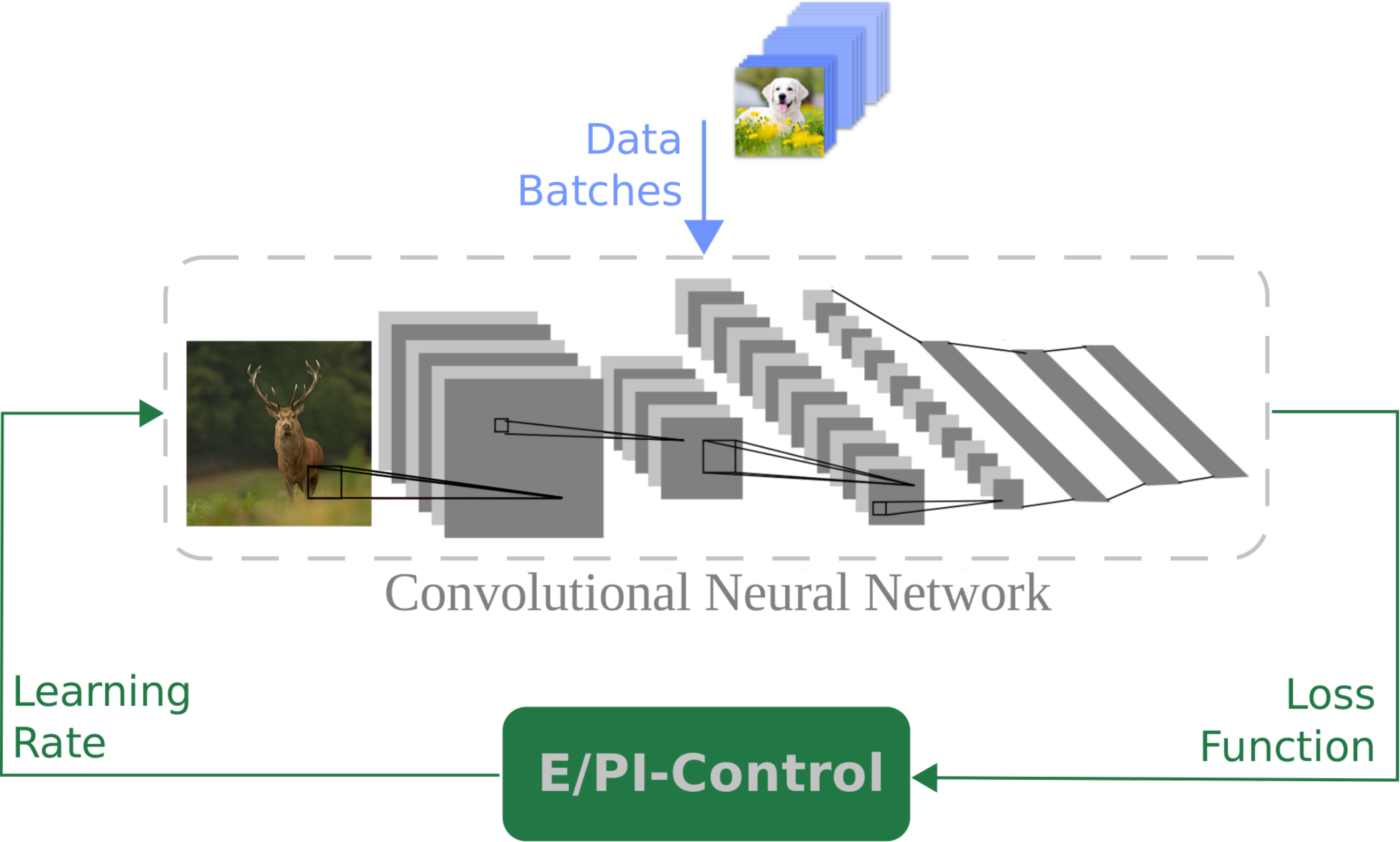}
		\caption{CNN control schema.}
		\vspace{-1em}
		\label{fig:schema}
	\end{center}
\end{figure}

We consider a CNN as being our plant, and the data used in training are seen as a disturbance, see Figure \ref{fig:schema}.
  
	\subsection{The disturbance: data batches}
We consider a training dataset that consists of several data instances, such as images. % Image data instance uses pixels' value as their features, so the number of features are according to their image size.
 And each data instance belongs to only one class $c$, where $c \in \{1, \dots, \textit{C}\}$, representing for instance the main object on the picture. Data instances, structured in batches, are assumed to arrive sequentially to the learning system over time. %, in the case of video batches or periodic data collections. 
%Due to the complexity of image data, each image is learned several times.
 We set that each batch consists of $B$ data instances. %, and will train several times on all data of the batch. 
One iteration on the whole batch is called an epoch, $E$ epochs are run on each batch. % During each epoch, we iteratively train our learning system with each data instance within the batch.
When the data of the new batch arrives, we discard the previous data and continue learning only on the new ones. This enables to reduce the storing space and processing time compared to keeping all the data. The control system sampling time is thus one epoch, while the disturbance time scale is the batches' one.

	\subsection{Performance Metrics: accuracy and loss}
	\label{sec:Performance_Metric}
%0
Two signals can be used to evaluate a neural network performance: validation accuracy and loss function, both varying through epochs.  
%1
Validation accuracy, computed on the test set, is the percentage of instances for which the predicted class matches the ground truth. 
%2
Loss function also compares the model predictions with the ground truth, but includes the notion of confidence in the prediction through the use of a distance. 
%2'
Cross-entropy is one of the most used multiclass classification loss function, and is thus the one we selected to be our output signal.
 It is the sum of cross-entropy error between targets and the predicted values done on the test set. We define $y_{i,c}$ as being the ground truth, indicating for each image $i$ if it belongs to class $c$ ($y_{i,c}=1$) or not ($y_{i,c}=0$). $\hat{y}_{i,c}$ is the CNN output, indicating the predicted probability of the image $i$ to belong to class $c$. 
The loss function thus defined as follows:
\begin{align}
loss=-\frac{1}{T}\sum\limits_{i=1}^T \sum\limits_{c=1}^C \ y_{i,c} log(\hat{y}_{i,c}) + (1-y_{i,c}) log(1-\hat{y}_{i,c})
\end{align}
with $T$ the size of the test set. The lower the value of loss function, the better the model is for the data, and as the entropy can't be negative, the target loss value is 0. Hence, the loss function is our error signal. 

Validation accuracy is %however
 used as an \textit{a posteriori} evaluation signal, as eventually, only the final prediction matters, whatever its confidence. Several metrics are extracted from the accuracy signal to reflect the CNN training performances. The end value of the accuracy is indeed the key factor. However, accuracy's converging speed is also important engineering concern: for complex image datasets such as ImageNet, state of the art training strategies take up to a few days. %Even with GPU or TPU support and improved algorithms, the training time is still counting by hours. So accuracy's converging time is vital for the large-scale dataset. 
Another important metric is the stability of the accuracy curve, especially for the epochs toward the end of the learning. %Indeed, as only the CNN parametrization of the last epoch will be used, 
A low standard deviation of the accuracy provides more guarantees on the final CNN performances.
%Because this is the period the accuracy should steadily converge to its ending point, otherwise, the accuracy will vary a lot between epochs. The strong oscillations are normally due to a big learning rate, it makes loss function can't precisely approach the optimum.
Eventually, the final value of the loss function is also of interest, as it allows to evaluate the model overfitting on data when compared to accuracy value.

	\subsection{The Control Signal: the learning rate}
\begin{figure}[!ht]
	\begin{center}
		\subfloat[Validation accuracy]
		{
			\includegraphics[width=0.47\columnwidth]{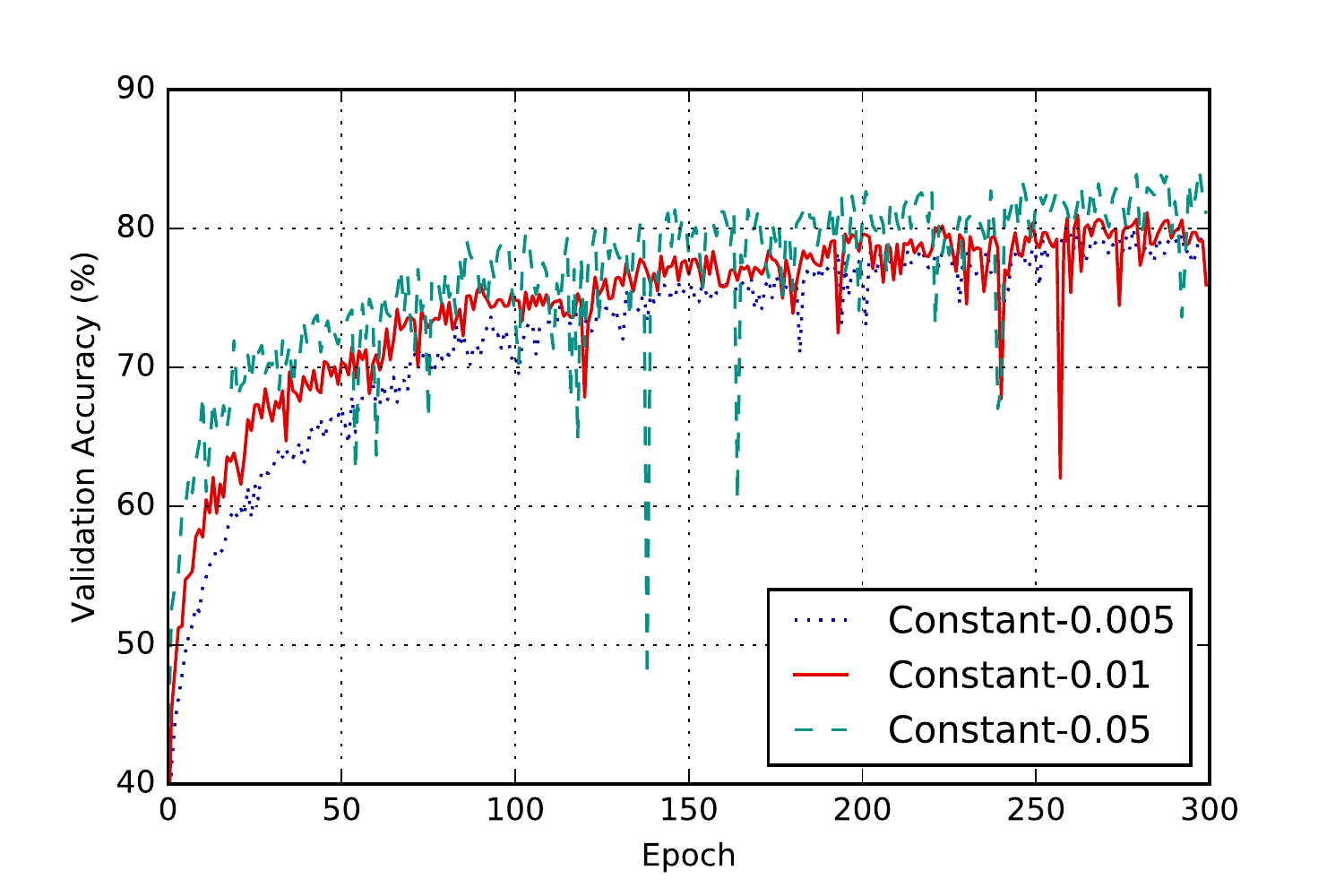}
		%	\vspace{-5em}
			\label{fig:Constant-accuracy}
		}
		\hfil
		\subfloat[Loss function]{
			\includegraphics[width=0.47\columnwidth]{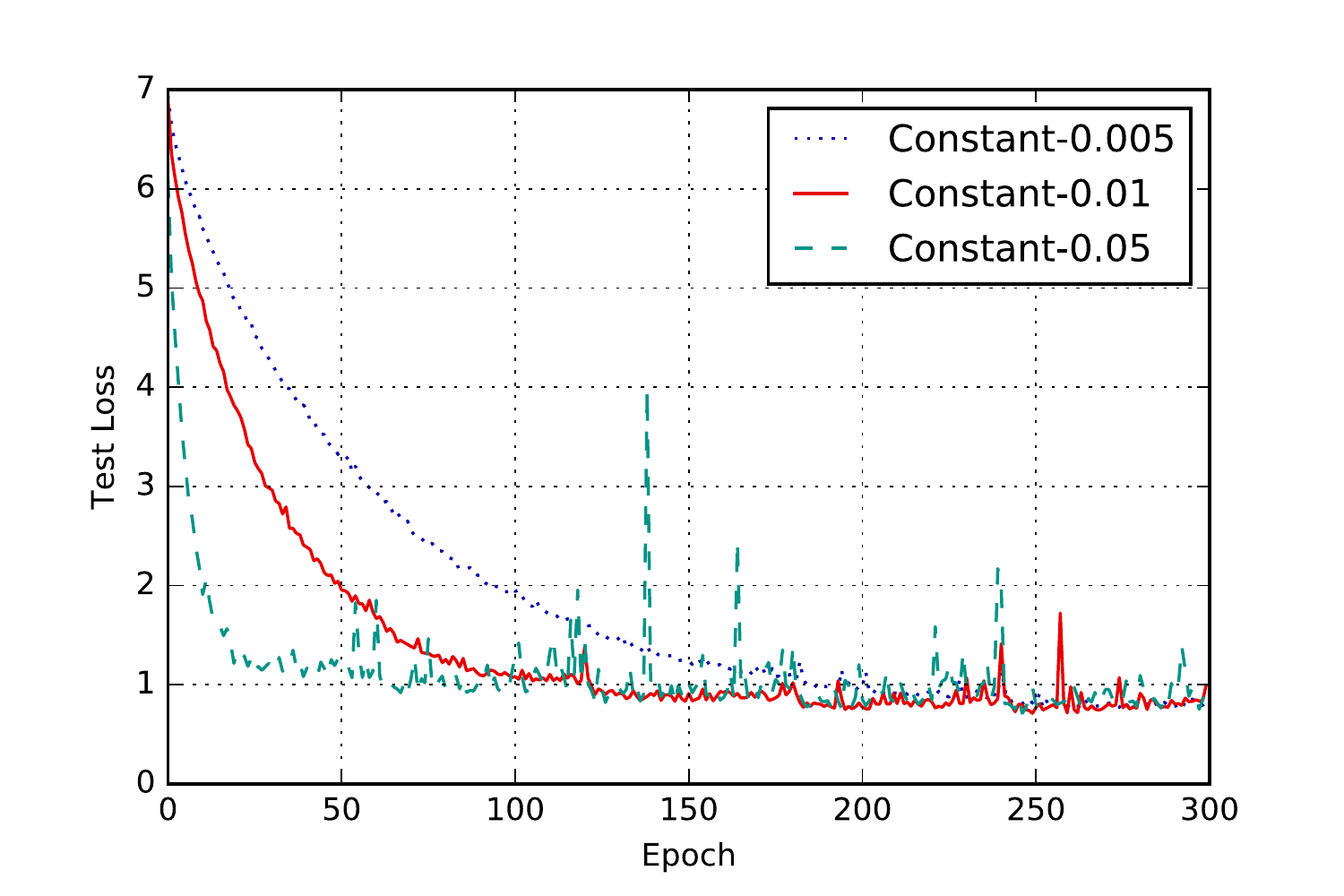}
	%		\vspace{1em}
			\label{fig:Constant-loss}
		}
		\caption{Impact of different constant learning rates on accuracy and loss (CIFAR-10).}
		\vspace{-2em}
		\label{fig:Constant-result}
	\end{center}
\end{figure}

The learning rate $\lambda$ is our control signal. %: it is a hyperparameter that controls how much we are adjusting the weights of our network with respect to the loss gradient.
To illustrate that the learning rate is a control signal for our online scenario, we study the impact of different constant learning rate on the variation of the accuracy and loss functions over epochs. Figure \ref{fig:Constant-result} is the application of three constant learning rates ($\lambda \in \{0.005,0.01,0.05\}$, corresponding to Bengio's recommendation \cite{Bengio2012}, larger and smaller) for training on CIFAR-10 dataset, with new data batches arriving every 60 epochs (see Section \ref{sec:Evaluation_setup} for more details). %We divide Cifar-10 (A 10 classes color image datasets) training dataset into 5 batches, each batch consists of 10000 images, we have another 10000 images as validation dataset to evaluate the accuracy and loss of our model at end of each epoch. Each batch of data will learn 60 epochs. Figure.~\ref{fig:Constant-result} shows the variation of validation accuracy and loss over time under different constant learning rates (of ). Each of the three experiments use the same learning algorithm, the only difference between the experiments is the learning rate. We can clearly observe that 
The accuracy and loss signals differ according to the learning rate (see Figure~\ref{fig:Constant-result}): with $\lambda=0.05$, the accuracy improves the fastest and the loss also quickly converges to its lower limit. However, the noise of the curve at the last epochs is also higher than with the two other scenarios because a large learning rate oscillates around the minimum. When $\lambda=0.005$, the accuracy increase is slower but the loss function varies more smoothly, the loss value rarely rises. 

Thus, the learning rate is able to influence our performance indicators: it is suitable as a control signal.

\section{Motivation}
\label{sec:Motivation}
The experiments shown in Figure \ref{fig:Constant-result} illustrate the advantages and drawbacks of large and small learning rates. A natural thought is to combine their benefits through learning rate scheduling: an initial phase with a large learning rate to quickly converge to a high-level accuracy, then a smaller value to smoothly approach the minimum and avoid the bumps on validation accuracy and loss.
In the state of the art, there are some commonly used learning rate strategies that vary the learning rate through time. %We set up an identical scenario as we did in section~\ref{sec:Motivation}, the only variable is the different learning rate stratety, we call them: (1) Step-decay learning rate strategy,
In the following of the paper, we will introduce two learning rate laws: (i) Keras-Time-Based-decay  and (ii) Exponential-Sine-Wave-decay. They will later be used in Section \ref{sec:Evaluation} to compare with our proposed methods. %The corresponding variation of validation accuracy, loss and learning rate are showed in Figure.~\ref{fig:Stateofart_result}. \sophie{say that we reset the strategy at each new batch.}

%Step-decay strategy decreases the learning rate to half of the learning rate before after each $S$ epochs, here in Figure.~\ref{fig:Stateofart_learning_rate}, $S = 10$.

Keras-Time-Based-decay is a commonly and widely used learning rate strategy in Keras\cite{chollet2015keras}, which is a famous python deep learning library. The learning rate is computed as follow:
\begin{equation} \label{eq:keras_time_based}
\lambda(k)=\dfrac{\lambda(k-1)}{ 1 + \delta k}
\end{equation}
where $k$ is the number of epochs since the arrival of the last batch, $k \in \{1, \dots, E\}$. $\delta$ is a hyperparameter enabling to tune the steepness of the time decay. We set $\lambda(0) = 0.01$ and $\delta = 0.001$ as suggested in \cite{Bengio2012}.

The second common schedule is the exponential decay, it has been successfully used in neural network training. A good implementation is  exponential decay sine wave learning rate schedule~\cite{W8305126}. The original schedule is implemented to an offline setting, so to adapt this learning rate schedule into our online setting, we need to adjust their strategy to allow the learning rate decays to around 0 at the ending epochs of each batch. We will refer to this strategy as Exponential-Sine-Wave-decay. The adapted version is calculated as follow:
\begin{align} \label{eq:exponential_sine_wave}
\lambda(k)= \lambda(0) e^{\frac{-\alpha k}{E}}(\gamma sin(\frac{\beta k}{2\pi}) + e^{\frac{-\alpha k}{E}} + 0.5)
\end{align}
where $k$ shares the same definition as in eq. \eqref{eq:keras_time_based}. $E$ is the training epochs per batch. $\alpha$, $\beta$ and $\gamma$ are three hyperparameters. In order to have a same behavior as in~\cite{W8305126} during our shorter $E$, we set $\alpha = 3$, $\beta = 6$ and $\gamma = 0.4$. The constant 0.5 in the equation is important, it makes sure that $\lambda(k)$ is strictly positive. %The corresponding learning rate schedule for training of CIFAR-10 dataset is showed in Figure \ref{fig:Stateofart_learning_rate}.

\section{Performance-based Learning Rate Laws}
\label{sec:Control}
In all the related work strategies, the learning rate is decreasing with time, in a predefined manner. The only differences between these strategies is that for some the learning rate decreases slowly in the beginning and faster in the end and for others is the opposite. % as it can be seen from Fig~\ref{fig:Stateofart_result}. %We think therefore that a feedback control to automatically change the learning rate based on the neural network performance (i.e. loss) is more suitable and robust. 
We therefore introduce our control strategy where the learning rate is automatically computed based on the loss function (see Fig~\ref{fig:schema}). Nevertheless, according to the definition of the loss function, the absolute loss value in itself does not give us much information since different size of training dataset can change the absolute loss value. Therefore, we normalize the value of $loss(k)$ by $loss(k=0)$, where $loss(k)$ represents the loss value at $k^{th}$ epoch since the arrival of the last batch.

Subsequently, we try three different control laws for computing the learning rate $\lambda$: Proportional-Control (P-Control), Proportional Derivative-Control (PD-Control) and a Mixed Exponential PD-Control (E/PD-Control).

\subsection{P-Control} In this case the learning rate depends proportionally on the loss value as follows:
\begin{align}
 \lambda(k) = K_P \dfrac{loss(k)}{loss(0)}
\end{align}
In general the value of $\frac{loss(k)}{loss(0)}$ varies between $0$ and $1$. Indeed, as the loss function decreases thanks to the Stochastic Gradient Descent, we know that we are approaching the minimum of the loss function and therefore the learning rate should be decreased in order not to skip it. %If however the value of $loss(k) / loss(0)$ is higher than one, it means we have skipped the minimum. %% --> jsutifies that we use a I action, so moved to the next subsection.
%Therefore, we also need a higher learning rate to decrease loss again, so the P-Control also makes sense.
The choice of $K_P$ is important for the speed of convergence. Based on trial and error tests, we make it equal to the same value as the empirical starting learning rate from \cite{chollet2015keras}: $\lambda(0)=K_P=0.01$.

\subsection{PD-Control} On one side, the hypothesis behind P-Control is that the loss is always decreasing; as we are getting closer to a minimum, the learning rate should slow down to better approach it. On the other side if the loss has decreased during last epoch 
%(i.e.$-(loss(k) - loss(k-1))$ is positive),
 we are in the good direction to find the minimum so we should reward last learning epoch by increasing the learning rate. This can be seen as adding an integral action to our controller. We express our PD-Control as follows:
\begin{align}
 \lambda(k) = K_P \dfrac{loss(k)}{loss(0)} - K_D \dfrac{loss(k) - loss(k-1)}{loss(0)}
\end{align}
where $K_P=0.01$, and the integral parameter is empirically chosen at 5 times $\lambda(0)$, as we choose $\lambda(0) = 0.01$ , then $K_D=0.05$. %In this control law 0.05 is a hyper-parameter chosen by us but we could also obtain this value by doing cross validation on a small part of dataset. 
As $-(loss(k) - loss(k-1))$ could also be negative, the integral part will introduce oscillations to the learning rate. In order to avoid that $\lambda(k)$ becomes negative due to the integral part, the PD-Control is turned to a P-Control if $\lambda(k)$ in PD-Control gets a negative value. Indeed, the P-Control will always return a positive value for the learning rate, as the loss function is by definition positive.
%Figure~\ref{fig:Control_p_pi} in Section~\ref{sec:Evaluation} shows the comparison of the learning rate of P and PI control. The difference in the convergence speed between two curves is subtle but we can definitely observe that the curve of PI Control's learning rate has more noise than P Control which is due to the I-Control part as we expected.\sophie{ The learning rate periodically returns to 0.01 is because when a new batch comes, we will force that the learning rate of epoch 0 is 0.01. So we will always use 0.01 as learning rate to welcome new batch, aims to avoid overfitting the first come data, and learn very little thing from latter come data. - let's move it in Section V next to the validation ?}
%% --> moved to evaluation.

\subsection{E/PD-Control} %\sophie{To present as a combination of a Exponetial law for the learning rate and a PI controller.}
This third control law tries to accelerate the convergence speed by exponentially increasing the learning rate at the beginning of learning a new batch, as the data are new so there are more informations to learn. %Following the idea behind the I-Control part, if the loss decreased during last epoch it means that we are in the good direction to find the minimum but we have not yet reached that point.  
We present a two phases algorithm to control the learning rate: (i) an initial Exponential growth followed by (ii) a PD-Control. During the exponential growth period, the learning rate is increased each time step by a factor 2 to quickly reach the minimum. This phase is stopped when the loss starts increasing, and the learning rate is afterwards ruled by the PD-Control law. The PD phase is initialized with the last value of the learning rate before loss growth. The PD parameters are set according to the behavior of the systems during the E-phase. The E/PD-Control law during one batch is summarized in Algorithm \ref{alg:EPI}. 

\begin{algorithm}
	\caption{E/PD-Control}
	\label{alg:EPI}
	\begin{algorithmic}[1]
%		\Function{$A$}{$\mathit{UD}$,$\mathit{KD}$} 
		\State $\lambda(0)=0.01$
		\State $k=1$
		\While {$loss(k) \leq loss(k-1)$}
		\State $\lambda(k)=2\lambda(k-1)=2^k\lambda(0)$ 
		\State $k=k+1$
		\EndWhile
		\State $\lambda(k+1)=\lambda(k)/2$
		\State $K_P=\lambda(k)/2$
		\State $K_D=5\times\lambda(0)$
		\For {$  i \in \{k+2,\dots ,E\}$}
		\State $\lambda(k)=K_P \dfrac{loss(k)}{loss(0)} - K_D \dfrac{loss(k) - loss(k-1)}{loss(0)}$
		\EndFor
	\end{algorithmic}
\end{algorithm}

%At beginning of each batch, if $(loss(k) - loss(k-1))<0$, the loss is decreasing, then:
% \begin{align*}
% \lambda(k+1) = 2 \times  \lambda(k) = 2^{k+1} \times  \lambda(0)
%\end{align*}
%once $(loss(k) - loss(k-1))>0$ we missed the minimum then we will cut half of the learning rate:  
%\begin{align*}
% \lambda(k+1) = 0.5 \times  \lambda(k)
%\end{align*}
%and from this moment we will let the PI-Control steadily approach the minimum. %The kp dynamically chosen by ending point of Fast Control phase is more representative than 0.01. 
%According to Figure~\ref{fig:Constant-result}, we know that 0.01 may not be the best starting learning rate so therefore the Fast Control law makes sense. 
%
%If the system is at its beginning or a we have a new batch which is contains useful informations Fast Control phase will last several epochs then the PI-Control phase will start but from a higher value. Even if a batch doesn't contain much information, even two or three epochs of Fast Control will be very helpful to increase the converge speed of the loss value to its steady state level.

\section{Control Laws Evaluation}
\label{sec:Evaluation}

In this section, the P, PI and E/PI-Control laws are evaluated in comparison with the state of the art. The stability of the E/PI-Control law is highlighted, and its robustness with regards to its initial configuration is presented. First, details on the datasets, CNNs and evaluation indicators are given.

	\subsection{Experimental setup}
	\label{sec:Evaluation_setup}
%datasets+CNN config:
The controllers are evaluated on two datasets: CIFAR-10 %(a natural image dataset with 10 categories) 
and Fashion-MNIST. %(a 10 classes grayscale image dataset of fashion articles). 
 The CNN and scenario configurations for the two datasets is sum-up in Table~\ref{tab:datasets}. As the images in CIFAR-10 have colors and are larger than the ones of Fashion-MNIST, a more complex CNN setting with more layers and parameters is used. Meanwhile, as there are more informations to extract from CIFAR-10, the number of epochs per batch is larger, allowing the accuracy curve to converge. All the values of hyperparameters of eq. \eqref{eq:keras_time_based} and \eqref{eq:exponential_sine_wave} we showed in section.\ref{sec:Motivation} are tuned for CIFAR-10, as Fashion-MNIST has shorter epochs per batch, we will change $\delta$ to 0.01 of eq. \eqref{eq:keras_time_based}, and set $\alpha = 2$, $\beta = 18$ of eq. \eqref{eq:exponential_sine_wave} for Fashion-MNIST experiment learning rate schedule.% in each batch. To remove the influence of mini-batch size to the learning process, we fix it to 128 for all the experiments. %% --> not necessary as we never talked about mini-batches before.

To eliminate the influence of the CNN's weights starting point to the final accuracy, we initialize the weights of each layer of CNN by Xavier uniform initializer \cite{DBLP:journals/jmlr/GlorotB10}, all the results will be averaged on 3 time experiment results. All code is implemented with Keras library\cite{chollet2015keras}.

\begin{table}[h]
	\begin{center}
		\caption{CNN configuration }
		\label{tab:datasets}
		\begin{tabular}{L{3.5cm} C{2cm} C{2cm}}
			\toprule
			\textbf{Use case}	& \textbf{CIFAR-10}	& \textbf{Fashion-MNIST} \\
			\midrule
			\#data instances to train	& 50,000					& 60,000 \\
			\#data instances to test T	& 10,000					& 10,000 \\
			\#classes $\textit{C}$ 		& 10 						& 10  \\
			image size & 32$\times$32 & 28$\times$28 \\
			data batch size $B$ 	& 10000					    & 10000   \\
			\#trainng epochs per batch $E$ 	& 60					    & 20   \\
			%mini-batch size in one epoch & 128 & 128 \\
			\#CNN layers	& 28					    & 15   \\
			\#CNN parameters 	& 1,641,858					    & 422,538 \\
			\bottomrule
		\end{tabular}
	\end{center}
\end{table}

%metrics:
For performance evaluation, we measure several indicators on the accuracy and loss signals, the first one being their final value. 
To quantify the accuracy's converging speed, we will report for each experiment the epoch at which they reached 95\% of their final accuracy. %This is our rising time. 
To compare the influence of each learning rate strategy on stability of validation accuracy, we will also compare the standard deviation of the accuracy curve on the last 10\% epochs of each experiment. % Only comparing the standard deviation of this period is reasonable because, these epochs are the time the accuracy should stabilize, steadily converge to the final accuracy. Otherwise, the model becomes unpredictable. %% --> already said before.

	\subsection{Stability analysis}
Stability of the presented algorithms needs to be proved to ensure that the error signal (the loss function) will not diverge, and ideally converge to 0. The stability theory behind the algorithm is based on the SGD: the direction of the gradient is always set to decrease the loss. The only case that loss will increase is because the learning rate is too large (i.e. we skipped the minimum). 
The stability of the P and PD-Controllers are ensured via a proper parametrization of $K_P$ and $K_D$. The E/PD-Control law allows the learning rate to exponentially grow, however the learning rate is switched to a PD law as soon as the loss increases. The reset of the learning rate to the previously stable value ($7^{th}-9^{th}$ line of Algorithm \ref{alg:EPI}) enables to properly initialize the PD.

% E -Control will stop when we firstly face the increase of loss, then enter to PI-Control. In general, the trend of learning rate during PI-Control period should be decrease. I-Control part could cause an augmentation of learning rate during one or two epcohs in PI-Control period, but that means we have a big descent on loss value, then P-Control part will sharply decrease. Even we encounter the situation that loss increases during the PI-Control period, I-Control part will decrease the learning rate, not allow it to increase too much. So learning rate of PI-Control will decrease over time with loss. The closer we approach the minima, the smaller the learing rate.
%One extreme case is that calculated learning rate during PI-Control is negative. This is because we have big jump in loss, so the I-Control part decreases learning rate too much. But in this case, we will turn to use P-Control, so we will avoid to reverse the direction of gradient.

\subsection{P, PD and E/PD-Control Performances Validation}

\begin{figure}
		\begin{center}
			\subfloat[Validation accuracy\vspace{-1em}]{
				\includegraphics[width=0.47\columnwidth]%{figures/cifar-10-control_accuracy_with_p}
				{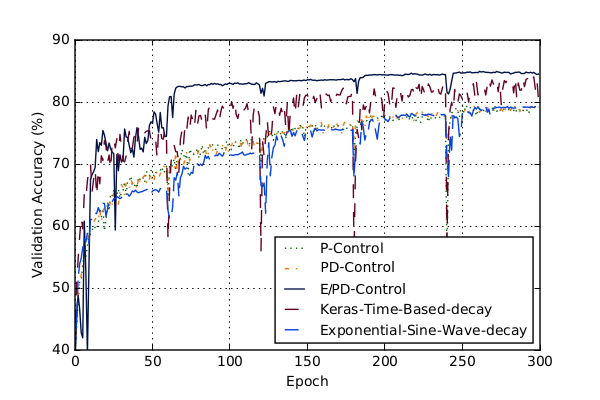}
				\label{fig:Control_accuracy}
			}
			\hfil
			\subfloat[Loss function]{
				\includegraphics[width=0.47\columnwidth]%{figures/cifar-10-control_loss_with_p}
				{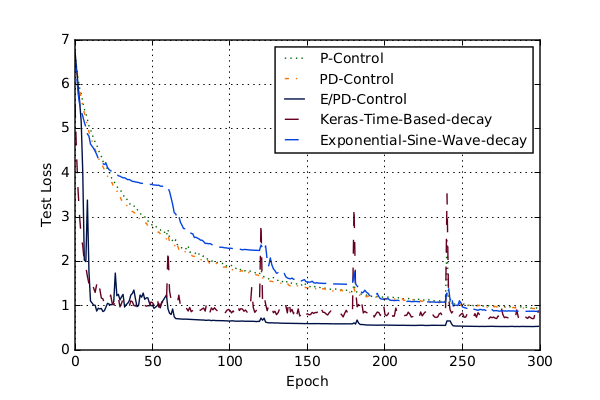}
				\label{fig:Control_loss}
			}
			
			\caption{Performances of state of the art, P, PD and E/PD-Control (CIFAR-10).}
			\vspace{-1em}
			\label{fig:Control_result}
		\end{center}
	\end{figure}

\begin{figure}
		\begin{center}
			\subfloat[P and PD-Control\vspace{-1em}]{
				\includegraphics[width=0.47\columnwidth]%{figures/cifar-10-Control_learning_rate_pi_with_p}
				{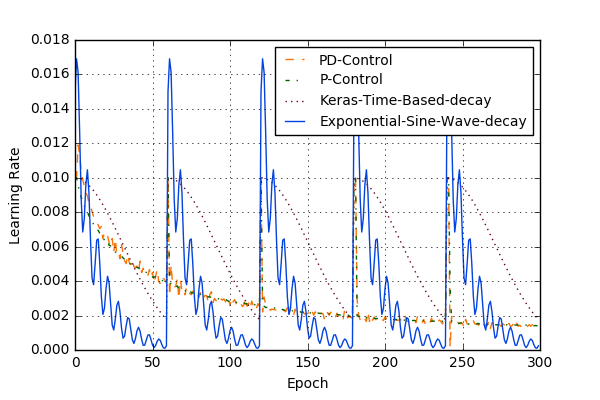}
				\label{fig:Control_p_pi}
			}
			\hfil
			\subfloat[E/PD-Control]{
				\includegraphics[width=0.47\columnwidth]{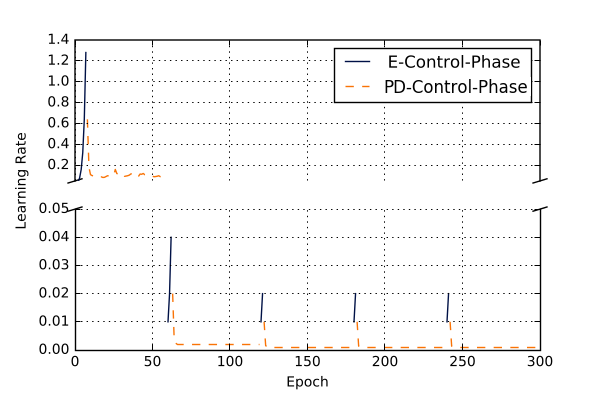}
				\label{fig:Control_fast}
			}
		
			\caption{Control signal of state of the art, P, PD and E/PD-Control (CIFAR-10).}
				\vspace{-1em}
			\label{fig:Control_ppi_fast}
		\end{center}
	\end{figure}

The three control laws presented in Section \ref{sec:Control} are evaluated on CIFAR-10. Results are reported in Figure \ref{fig:Control_result} through the accuracy \ref{fig:Control_accuracy} and loss functions \ref{fig:Control_loss}, and the corresponding control signals are illustrated in Figure \ref{fig:Control_ppi_fast}, For the P and PD in \ref{fig:Control_p_pi} and for the E/PD law in \ref{fig:Control_fast}. There are few differences between P and PD control performances, while the E/PD-Control is significantly faster (61 epochs rising time compared to 130 for the P and PD), converges to a higher accuracy (+7\%) and lower loss (-37\%) and the standard deviation of the accuracy at the end of the experiment is three times lower. We see from the first epochs that the E-phase enables to properly tune the initial value of the PD, which then significantly increases the validation accuracy.  
%From Table.\ref{tab:results_control_cifar10}, we can see that . Figure.\ref{fig:Control_p_pi} shows that at the beginning, as the loss decreases fast, I-Control plays an important role, and we can see the learning rate of PI-Control is slightly bigger than P-Control. This results in Figure.\ref{fig:Control_loss}, 
The P and PD-Control learning rate signal (Figure \ref{fig:Control_p_pi}) illustrates that a reset of the learning rate at the arrival of a new batch is not necessary beneficial if the value is not carefully chosen, as for the E/PD-Control.

The loss function with the PD-Control declines a little bit faster than with the P-Control at beginning, and do not present a large peak around epoch 240. Those advantages made us opt for the PD-Control to combine with the initial E-phase.

%Figure.\ref{fig:Control_result} and  Figure.\ref{fig:Control_fast} show that at the first batch, as our loss is far from optimum, the loss continues to sharply decline as we increase the learning rate exponentially. Even though there are big vibrations on the curve of  accuracy, we know that loss is a more stable measurement for the model, and the loss quickly converges and stabilizes to a low level. And in the following epochs, the stability of curve of accuracy and loss are all very good, it steadily converges to the better level. Compare the results in Table.\ref{tab:results_control_cifar10}, E/PI-Control are better in all the indicators. The best results are highlighted on \textbf{bold}.

	\subsection{Comparison with state of the art}

\begin{figure}[h]
	\begin{center}
		\subfloat[Validation accuracy \vspace{-1em}]{
			\includegraphics[width=0.47\columnwidth]{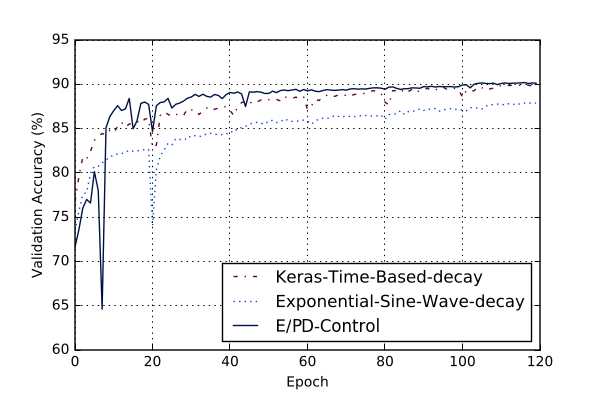}
			\label{fig:Fashion_mnist_accuracy}
		}
		\hfil
		\subfloat[Loss function]{
			\includegraphics[width=0.47\columnwidth]{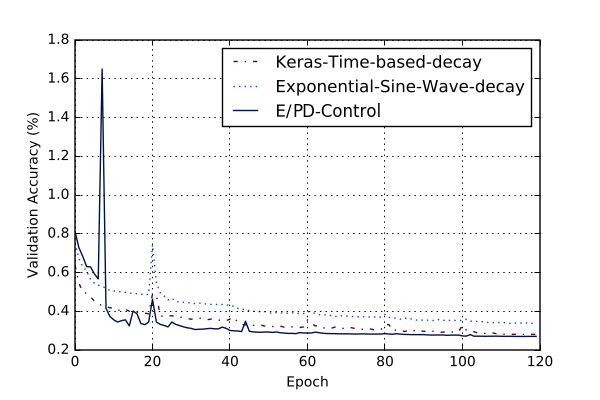}
			\label{fig:Fashion_mnist_loss}
		}
		\caption{Performances of state of the art and E/PD-Control on Fashion-MNIST.}
		\vspace{-1em}
		\label{fig:Fashion_mnist_result}
	\end{center}
\end{figure}

\begin{figure}
	\begin{center}
		\subfloat[State of the art strategies\vspace{-1em}]{
			\includegraphics[width=0.47\columnwidth]{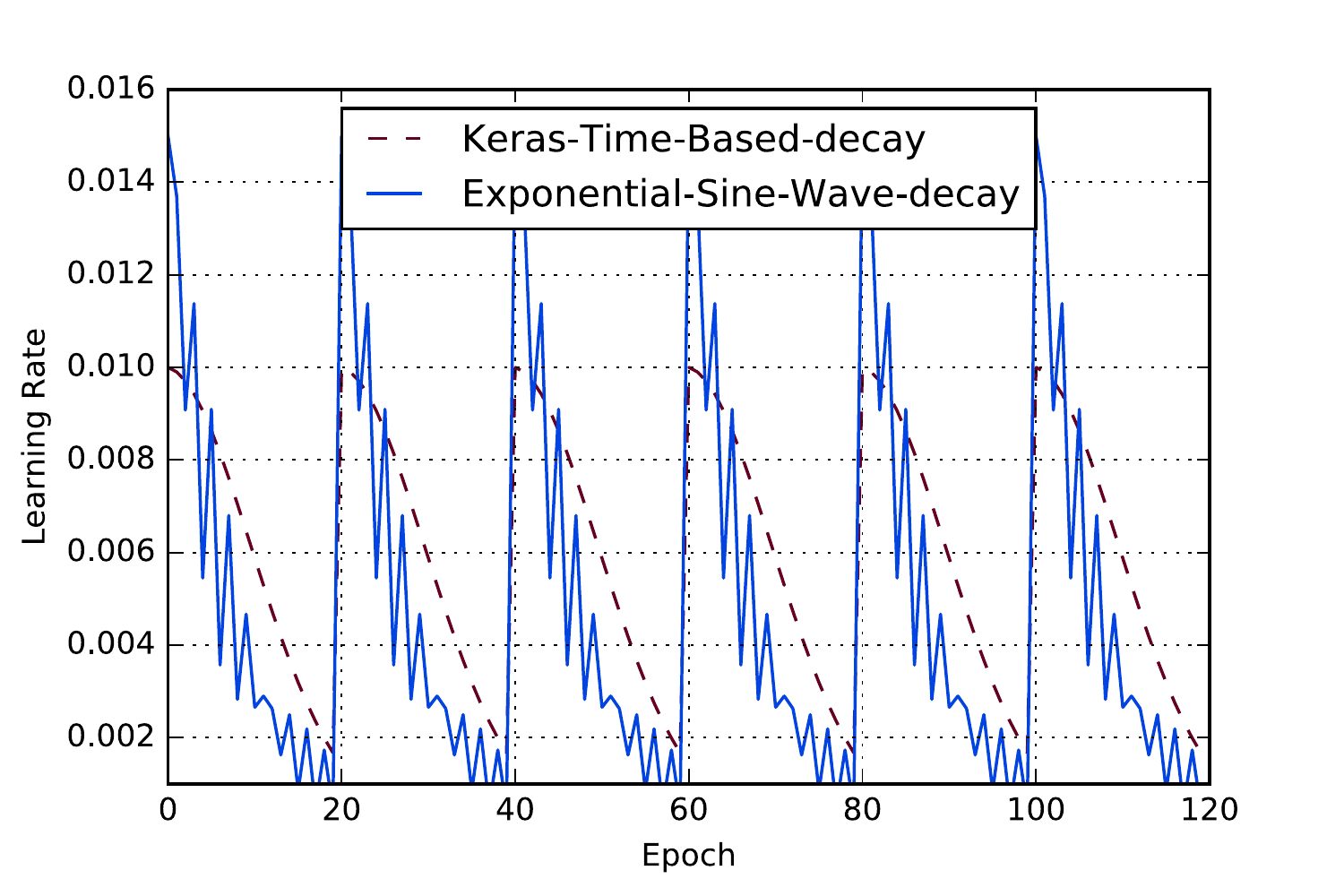}
			\label{fig:Fashion_mnist_empirical_learning_rate}
		}
		\hfil
		\subfloat[E/PD-Control]{
			\includegraphics[width=0.47\columnwidth]{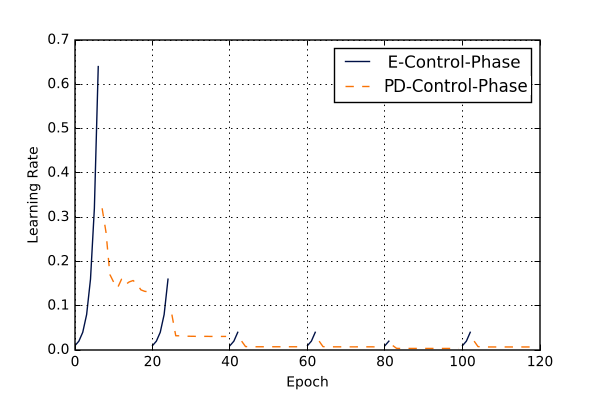}
			\label{fig:Fashion_mnist_fast_pi_learning_rate}
		}
		\caption{Control signal for the state of the art and E/PD-Control on Fashion-MNIST.}
		\vspace{-1em}
		\label{fig:Fashion_mnist_learnig_rate}
	\end{center}
\end{figure}

Comparison of the state of the art learning rate strategies to our E/PD-Controller is provided for CIFAR-10 (Figures %\ref{fig:Stateofart_result},
 \ref{fig:Control_result} and \ref{fig:Control_ppi_fast}) and for Fashion-MNIST (see Figure \ref{fig:Fashion_mnist_result} for the accuracy and loss and Figure \ref{fig:Fashion_mnist_learnig_rate} for the learning rate evolution through epochs).

E/PD-Control provides the best results for all the indicators for CIFAR-10. It converges faster and has a smallest standard deviation of last 10\% epochs among all the strategies, it reaches at a higher final validation accuracy (+3\%) and a lower loss. Keras-Time-Based-decay has a closer final accuracy and loss to E/PD-Control. But the deviation of its accuracy curve is bigger than E/PD-Control, especially at the beginning of learning a new batch.

Regarding Fashion-MNIST dataset, results are similarly in favor of the E/PD-Controller, even if the differences are smaller. As this dataset is easier than CIFAR-10, all strategies reached a high validation accuracy and lower loss, the standard deviation of accuracy of last 10\% epochs is also very small. %But still, we can see E/PI-Control has a better performance in all indicators. 
%It only takes 9 epochs to firstly reach 95\% of final accuracy. It stabilizes after No.50 epoch, and maintains to be the best accuracy in the last 70 epochs.
%E/PI-Control shows a drop in the accuracy when the law switches from E to PI phase, but except for the first one, the peaks are smaller than the ones of the state of the art strategies when a new batch comes in.

%Initial learning rate is still 0.01. To force the learning rate to decrease during 20 epochs, we adjust the hyperparameter $decay$ in Keras-Time-Based-decay to 0.01, the corresponding variations of learning rate are showed in Figure.~\ref{fig:Fashion_mnist_learnig_rate}, the variations of validation accuracy and loss are showed in Figure.~\ref{fig:Fashion_mnist_result}. The results are showed in Table.\ref{tab:results_Fashion_MNIST}. 

	\subsection{Robustness to initial value of the learning rate}
The E/PD-Control is now compared with the best strategy from the state of the art (Keras-Time-Based-decay) when the initial learning rate varies. %in order to check if the performance of our control strategy performs the best. %In order to find a target to compare, at each staring learning rate, besides E/PI-Control experiment, we will also run another experiment of Keras-Time-Based-decay, as it performs best among three empirical decay strategies. 
Results are showed in Table \ref{tab:results_Cifar_10} for CIFAR-10 and Table \ref{tab:results_Fashion_MNIST} for Fashion-MNIST. %Each experiment is done 3 times, both average and standard deviation are provided. The best results of each initial learning rate are highlighted in \textbf{bold}.

Among all the experiments on CIFAR-10, there are only one case for which Keras-Time-Based-decay law has a better indicator (final validation accuracy at initial learning rate 0.05). The difference is very small, and standard deviation of the indicator itself is large. Moreover, if we check the accuracy's standard deviation during the last 10\% epochs, Keras-Time-Based-decay still has a strong oscillation, which makes the model unpredictable. E/PD-Control also shows the advantage on converging time, it makes the model converge faster and rarely affected by the initial values.

%The final loss of E/PI-Control are all lower than Keras-Time-Based-decay. E/PI-Control's accuracy are very stable at the last 10\% epochs, its converging times are also extremely stable, it's all around 62 out of the 300 epochs. Another important point to notice is that, as initial learning rate varies from 0.001 to 0.1, accuracy of Keras-Time-Based-decay varies from 79\% to 85.6\%, while E/PI varies only from 82\% to 85\%. Thanks to the E-phase, even with a small initial learning rate, the E/PI-Controller gives good and fast results. This shows that Keras-Time-Based-decay is more sensible to the initial learning rate.

Table \ref{tab:results_Fashion_MNIST} shows the robustness results on Fashion-MNIST. E/PD-Control still shows a fast converging speed, reaching 95\% of final accuracy just using 7 to 10 epochs. The performances for the final loss and accuracy final standard deviation are similar for the two strategies.
%Even though here are not big differences on the indicators of Accuracy's standard deviation of last 10\% epochs and final loss, E/PI-Control still shows better performances in most of indicators. 
The E/PD-Control's final accuracy performances among all the experiments is more stable than with Keras-Time-Based-decay, which again, shows that E/PD-Control is more robust to the initial learning rate variations.

%To summarize  Table.\ref{tab:results_Cifar_10}, Table.\ref{tab:results_Cifar_10}, E/PI-Control could increase his final validation accuracy with the increasing of initial learning rate as good as Keras-Time-Based-decay, and it can still remain its stability and efficiency. When initial learning rate decreases, it can suppress the decline of the final validation accuracy not as fast as Keras-Time-Based-decay. Moreover, the converging speed of E/PI is not only fast, but also stable.

\begin{table*}[t]
	\begin{center}
		\caption{Robustness experiments with varying initial learning rate on CIFAR-10. Mean value (and standard deviation) over 3 runs. The best results are highlighted in \textbf{bold}.}
		\label{tab:results_Cifar_10}
		\begin{tabular}{L{1.5cm} C{2cm} C{1.5cm} C{3cm} C{3cm} C{3cm}}
			\toprule
			\textbf{Algorithm}	& \textbf{Initial learning rate}	& \textbf{Final loss} &\textbf{Final validation accuracy (\%)}	& \textbf{Final accuracy standard deviation} & \textbf{First epoch to reach 95\% accuracy} \\
			\midrule
			Keras	&0.001& 0.849(0.023)& 79.075(0.485)			&0.371(0.053)   & 161.333(24.495)/300\\
			E/PD-Control	&0.001& \textbf{0.648(0.015)}& \textbf{82.035(0.465)}			& \textbf{0.057(0.013)}  & \textbf{61.333(0.471)/300} \\
			\midrule
			Keras		&0.002& 0.745(0.026)& 80.180(0.445)		    & 0.415(0.049)  &118.333(9.78)/300\\
			E/PD-Control	&0.002& \textbf{0.586(0.006)}& \textbf{83.150(0.225)}			&\textbf{0.077(0.017)}   & \textbf{62(0)/300}\\
			\midrule
			Keras	&0.05& 0.727(0.006)& \textbf{85.640(0.523)}	& 1.630(0.085)  & 103.333(2.859)/300 \\
			E/PD-Control 	&0.05&\textbf{0.555(0.005)}& 85.060(0.090)			            & \textbf{0.117(0.001)}  & \textbf{61.333(0.471)/300} \\
			\midrule
			Keras	&0.1& 0.829(0.180)& 84.433(2.82)		&1.609(0.432)   & 77.333(32.785)/300\\
			E/PD-Control	&0.1& \textbf{0.578(0.013)}& \textbf{85.075(0.585)}					    &\textbf{0.345(0.16)}   & \textbf{65(0.816)/300}\\
			\bottomrule
		\end{tabular}
	\end{center}
\end{table*}

%%%%%%%%%%%%%%%%%%%%%%%%%%%%%%%%%%%%%%%%%%%%%%%%%%%%%%%%%%%%%%%%%%%%%
\begin{table*}[t]
	\begin{center}
		\caption{Robustness experiments with varying initial learning rate on Fashion-MNIST. Mean value (and standard deviation) over 3 runs. The best results are highlighted in \textbf{bold}.}
		\label{tab:results_Fashion_MNIST}
		\begin{tabular}{L{1.5cm} C{2cm} C{1.5cm} C{3cm} C{3cm} C{3cm}}
			\toprule
			\textbf{Algorithm}	& \textbf{Initial learning rate}& \textbf{Final loss}	&\textbf{Final validation accuracy (\%)}	& \textbf{Final accuracy standard deviation} & \textbf{First epoch to reach 95\% accuracy} \\
			\midrule
			Keras &0.001& 0.413(0)& 85.055(0.035)	&0.054(0)   & 37(0.816)/120\\
			E/PD-Control	&0.001& \textbf{0.334(0.002) }&\textbf{ 87.955(0.105)}			& \textbf{0.023(0.008)}  & \textbf{10.667(1.247)/120}\\
			\midrule
			Keras  		&0.002& 0.360(0.001)& 86.850(0.005)		    & 0.066(0.002)  &25.667(1.247)/120\\
			E/PD-Control	&0.002& \textbf{0.350(0.008)}& \textbf{87.415(0.103)}					&\textbf{0.057(0.011)}   & \textbf{8.333(0.471)/120}\\
			\midrule
			Keras 	&0.05& 0.282(0.026)& 89.785(0.920)			& 0.145(0.012) & 16.667(6.532)/120 \\
			E/PD-Control 	&0.05& \textbf{0.263(0.006)}& \textbf{90.425(0.200)}			& \textbf{0.094(0.013)}  & \textbf{9.333(1.700)/120}\\
			\midrule
			Keras 	&0.1& 0.265(0.016)& 90.400(0.674)					&0.133(0.010)   & 9(3.265)/120\\
			E/PD-Control	&0.1& \textbf{0.249(0.003)}& \textbf{91.340(0.140)}					&\textbf{0.114(0.015)}   & \textbf{7(0)/120}\\
			\bottomrule
		\end{tabular}
	\end{center}
\end{table*}

\section{Conclusion}
When performing image classification tasks with neural networks, often comes the issue of on-line training, from sequential batches of data. Iterative training of CNNs is driven by a learning rate - how much to update the network weights with the new data - which value is usually ruled by a time decreasing function. 
This paper presents a control approach to the challenge of on-line training of CNNs, that decides the learning rate value based on the expected learning need (i.e. the CNN loss function) instead of being time-based. 
E/PD-Control is a strategy that combines a phase of exponential growth of the control signal (i.e. learning rate) with a PD controller, which parameters are automatically adapted based on the E-phase. 

Stability of the control strategy is provided, and evaluation highlights that E/PD-Control achieves a higher accuracy level in a shorter time than the state of the art solutions. Robustness of the approach is illustrated by its performances on two different datasets, and enforced by a sensitivity analysis regarding its initialization.

This work could be further extended by the addition of a triggering mechanism to smartly adapt the number of epochs needed at each batch processing. Moreover, we want to investigate the performances of the E/PD-Control in the scenario when new classes appear in some batches.%See the impact of noise.

%\addtolength{\textheight}{-12cm} 

%%%%%%%%%%%%%%%%%%%%%%%%%%%%%%%%%%%%%%%%%%%%%%%%%%%%%%%%%%%%%%%%%%%%%%%%%%%%%%%%
%\section*{Acknowledgement}

\bibliographystyle{IEEEtran}
\bibliography{CCTA19}

\end{document}